%% file: main.tex
\documentclass[11pt,a4paper]{article}
\usepackage{geometry}
\usepackage[utf8]{inputenc} 
\usepackage{hyperref}       
\usepackage{url}            
\usepackage{booktabs}       
\usepackage{amsfonts}       
\usepackage{nicefrac}       
\usepackage{microtype}      

\usepackage{graphicx}
\usepackage{subcaption}  
\usepackage{enumerate}
\usepackage{paralist}
\usepackage{xcolor} 
\usepackage[hypcap]{caption}
\usepackage[ruled,vlined]{algorithm2e}

\usepackage{amsthm}
\usepackage{amsfonts}
\usepackage{amssymb}
\usepackage{mathrsfs}  
\usepackage{mathtools}
\usepackage{bm} 
\usepackage{bbm} 
\usepackage{float}
\newcommand{\mc}{\mathcal}

\newcommand{\N}{ \ensuremath{\mathbb{N}}}  
\newcommand{\Z}{ \ensuremath{\mathbb{Z}}}  

\newcommand{\R}{ \ensuremath{\mathbb{R}}}  
\newcommand{\C}{ \ensuremath{\mathbb{C}}}  
\newcommand{\T}{\ensuremath{\mathbb{T}}}

\DeclarePairedDelimiterX\set[1]{\lbrace}{\rbrace}{  #1 }  

\DeclarePairedDelimiterX\norm[1]{\lVert}{\rVert}{#1}  			
\DeclarePairedDelimiterX\inner[2]{\langle}{\rangle}{#1 \,,\, #2}  	
\newcommand{\mat}[1]{\mathbf{#1}}

\DeclarePairedDelimiterX\abs[1]{\lvert}{\rvert}{#1} 

\DeclareMathOperator*{\argmin}{arg\; min}

\usepackage{array, makecell}
\usepackage{wrapfig}

\title{Applications of Koopman Mode Analysis to Neural Networks}

\author{%
  Iva Manojlovi\'{c}\thanks{ 
  AIMdyn, Inc.
  Santa Barbara, CA 93101. (\texttt{imanojlovic@aimdyn.com}, \texttt{mfonoberova@aimdyn.com}, \texttt{mohrr@aimdyn.com}, \texttt{aleksandr@aimdyn.com}, \texttt{mezici@aimdyn.com})}
\and
  Maria Fonoberova\footnotemark[1]
\and
  Ryan Mohr\footnotemark[1]
\and 
Aleksandr Andrej\v{c}uk\footnotemark[1] 
\and
  Zlatko Drma\v{c}\thanks{
  University of Zagreb 
  10000 Zagreb, Croatia. 
  \texttt{drmac@math.hr}}
\and
  Yannis Kevrekidis\thanks{ 
  Johns Hopkins University
  Baltimore, MD 21218.
  \texttt{yannisk@jhu.edu}}  
\and
  Igor Mezi\'{c} \footnotemark[1]
}

\begin{document}

\maketitle

\begin{abstract}
  We consider the training process of a neural network as a dynamical system acting on the high-dimensional weight space. Each epoch is an application of the map induced by the optimization algorithm and the loss function. Using this induced map, we can apply observables on the weight space and measure their evolution. The evolution of the observables are given by the Koopman operator associated with the induced dynamical system. We use the spectrum and modes of the Koopman operator to realize the above objectives. Our methods can help to, a priori, determine the network depth; determine if we have a bad initialization of the network weights, allowing a restart before training too long; speeding up the training time. Additionally, our methods help enable noise rejection and improve robustness. We show how the Koopman spectrum can be used to determine the number of layers required for the architecture. Additionally, we show how we can elucidate the convergence versus non-convergence of the training process by monitoring the spectrum, in particular, how the existence of eigenvalues clustering around 1 determines when to terminate the learning process. We also show how using Koopman modes we can selectively prune the network to speed up the training procedure. Finally, we show that incorporating loss functions based on negative Sobolev norms can allow for the reconstruction of a multi-scale signal polluted by very large amounts of noise.
\end{abstract}

\section{Introduction}

The training of neural networks is a topic of much interest due to their wide spread use in a variety of application domains, from image recognition and classification to solving ordinary and partial differential equations. Unfortunately, the dimensionality of the problem often prevents rigorous analysis. Viewing a neural network training as dynamical system \cite{Chang:2017te,Kevrekidis:2020koopmanalgorithms, Chang:2019wk} provides a framework for a mathematical approach to training.

Our objective is to introduce the wider community to various results in applying methods from dynamical systems, in particular the operator-theoretic approach, to extract insight into both the choosing of the neural network's architecture, such as its depth and the pruning of weights, for a given problem and insight into the networks training process.

For many problems, massive amounts of compute power is thrown at the problem. When choosing network size, often a massive network is trained until good performance is obtained and then the network is iteratively trimmed while monitoring that the performance only degrades by an acceptable amount. This results in a smaller network that can be deployed with comparable performance. These methods do not tell us how to choose the network size a priori with respect to how many layers and nodes per layer. 

We consider the training process as a dynamical system acting on the high-dimensional weight space. Each epoch is an application of the map induced by the optimization algorithm and the loss function. Using this induced map, we can apply observables on the weight space and measure their evolution. The evolution of the observable is given by a linear operator, the Koopman operator, associated with the induced dynamical system. We use the spectrum and modes of the Koopman operator to obtain insight into the training process. Our methods can help to \emph{a priori} determine the network depth; determine if we have a bad initialization of the network weights, allowing a restart before training too long; pruning the weights of the network; and allow significant noise rejection when training on noisy multimodal signals

The rest of the paper is structured as follows. In section \ref{sec:nn-dyn-sys}, we mathematically formulate the training of a neural network as a dynamical system and introduce the Koopman operator viewpoint of analyzing dynamical systems. In section \ref{sec:training-convergence-koopman-spectrum}, we investigate the convergence of the training process through the lens of the Koopman operator's spectrum. In section \ref{sec:pruning}, we use the Koopman modes obtained via applying DMD analysis to the cross-entropy loss to pruning the weight of the network while maintaining performance. In section \ref{sec:hsvr}, we show how one can use either the Fourier or Koopman spectrum to determine the number of layers required to model a multiscale signal with a Hierarchical SVR model. The same ideas can carry over to using neural networks to model multiscale figures. In section \ref{sec:sobolev}, we apply loss functions inspired by negative Sobolev norms and show how they can be used for significant noise rejection when trying to reconstruct a signal.


\section{Neural network training as a dynamical system.}\label{sec:nn-dyn-sys}
Let $n(\mat x; \mat w)$, $n: X \times \R^n \to \R^d$, be a neural network, where $\mat x \in X \subset \R^m$ is the input feature vector,  $\mat w \in \R^n$ is the vector of network weights (parameters), and the output of the network is a $d$-dimensional real vector. Let $L_{tr}(\mat w)$ be the loss function of the network on the training set as a function of the network parameter weights. The optimization problem that is to be solved is
	\begin{equation}
	\mat w^* = \argmin L_{tr}(\mat w).
	\end{equation}

The loss function $L_{tr}$ and the chosen optimization algorithm (e.g. stochastic gradient descent) induce a nonlinear, discrete time map on the network weights that update the weights each epoch:
	\begin{align}
	&T : \R^n \to \R^n, 
	&\mat w_{t+1} = T(\mat w_t),
	\end{align}
where $\mat w_t$ are the network weights at the beginning of epoch $t \in \N_0$; $\mat w_0$ represents the initialized network weights prior to training. This induced map $T$ is a discrete dynamical system having the weight space as its state space.

Trying to directly analyze the map $T$ can be quite difficult as it is implemented as a black box in whatever training framework one is using. Instead one can track the evolution of observables to gain insight into the dynamical system by studying the spectral properties of an induced linear operator that drives the evolution of the observable.
Let $\mc F$ be a function space that is closed under composition with $T$; that is, if $f \in \mc F$, then $f \circ T \in \mc F$. An operator $U : \mc F \to \mc F$ can be defined via this composition $Uf = f \circ T$.
This operator, called the Koopman or composition operator, is linear \cite{Mezic:2005ug}, albeit infinite-dimensional, even if $T$ is nonlinear. Even though it is linear, it can still capture the full nonlinear behavior of $T$. In many cases, the Koopman operator has a spectral decomposition \cite{Mezic:2005ug,Budisic:2012cf,Mohr:2014wm,mezic2019spectrum}
	\begin{equation}\label{eq:koopman-spectral-decomp}
	U f = \sum_{j=1}^{\infty} c_j \lambda_j \phi_j + \int_\C dE(z)f
	\end{equation}
where $c_j \in \C$ is a coefficient, $\lambda_j \in \C$ is an eigenvalue, $\phi_j \in \mc F$ is an eigenfunction of $U$, and $dE(z)$ is a projection-valued measure (PVM). The set of eigenvalues form the point spectrum. The PVM is associated with the continuous spectrum of $U$; it takes sets in the complex plane and associates projection operators on $\mc F$ with them. We can also consider the case of vector-valued observables $\mat f = (f_1,\dots, f_m)$, where $f_i \in \mc F$. In this case, there is an analogous decomposition to \eqref{eq:koopman-spectral-decomp}, whose only difference is that the scalar-valued coefficients $c_j$ become vector-valued coefficients $\mat m_j \in \C^m$. These vector-valued coefficients are called Koopman modes \cite{Mezic:2005ug}. Data-driven algorithms, like the family of Dynamic Mode Decomposition algorithms (e.g. \cite{Schmid:2008wv,Rowley:2009ez,Jovanovic:2012wy,Jovanovic:2014ft,Hemati:2014jm,Williams:2015kh,Drmac:2018eka,Drmac:2018vpa}), are used to approximate the modes and eigenvalues using a trajectory from a single initial condition. Thus we do not need explicit access to $U$ to analyze the dynamical system. We use the \textsf{DMD\_RRR} algorithm from \cite{Drmac:2018eka} in the following work. For the rest of the paper we will assume that the observables we take will be contained in the subspace spanned by eigenfunctions and therefore their evolution (convergence or divergence) is controlled solely by the eigenvalues.

\section{Insights into neural network training convergence via the Koopman spectrum.}\label{sec:training-convergence-koopman-spectrum}

The main idea here is that by monitoring the spectrum of the Koopman operator during the training process will give us a method to determine when the training process should be terminated so that good performance is given on the testing set without the network memorizing the set. Having this indicator allow the network to generalize better.

Given the trajectory of all network weights $\set{\mat w_t}$, where $t\in\N$ is the training epoch, we use two different observables, a delay-embedded version of the (cross-entropy) loss function, $L_{tr}(\mat w_t)$ and the full-state observable which returns the network weights $\mat w_t$ at each training epoch $t$. Inspecting the Koopman spectrum of either of these observables tells us how quickly the system is learning and when a fixed point is appearing. The Koopman mode decomposition \eqref{eq:koopman-spectral-decomp} can be written as
	\begin{equation}
	L_{tr}(\mat w_t) = \sum_{k} \mat m_k \lambda_k^t \phi_k  + \mat e_t
	\end{equation}
where $\mat m_k$ is the Koopman mode normalized to norm 1, $\lambda_k$ is the associated eigenvalue, $\phi_k$ is the reconstruction coefficient and $\mat e_t$ is the error term at epoch $t$. If all the eigenvalues not equal to 1 satisfy $\abs{\lambda_k} < 1$, then the training is stable \cite{Mauroy2016:cf}. In this case, the slowest eigenvalue determines the rate of convergence. The importance of a mode is determined by the absolute value of the reconstruction coefficient, with higher values giving the mode more importance. The further inside the unit circle the eigenvalues corresponding to the important modes are, the faster the training.

\begin{figure}[htbp]
\begin{center}
\includegraphics[width=0.9\textwidth]{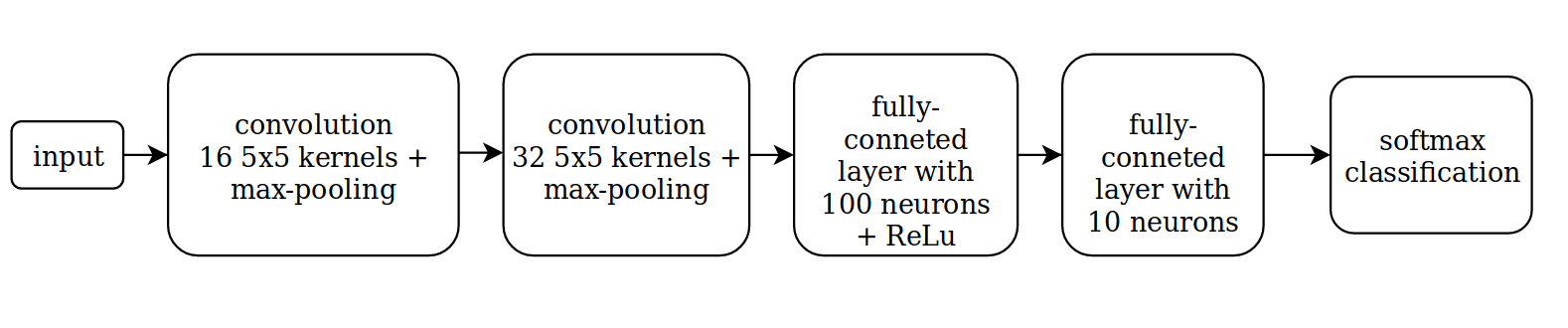}
\caption{Neural network architecture.}
\label{fig:small-nn}
\end{center}
\end{figure}

The standard MNIST data set was used for testing the methods. The network tested was a convolutional network with two convolutional layers with $16$ and $32$ kernels, each followed by $2 \times 2$ max-pooling. Each kernel is $5 \times 5$. Convolutional layers are followed by fully-connected layer with $100$ neurons, which is then followed by another fully-connected layer with $10$ neurons, after which follows softmax classification. The architecture is shown on Figure \ref{fig:small-nn}. A cross-entropy loss function was used with a learning rate of $\eta = 1e$-3. Different weight initialization (He or Xavier) were tested as well. After each epoch, cross-entropy loss on train and test sets were recorded along with the network weights. All networks were trained for $1000$ epochs, and KMD analysis was applied to the snapshots for 3 different epoch ranges specified in the figures.

\begin{figure}[htbp]
\begin{center}
\includegraphics[width=0.49\textwidth]{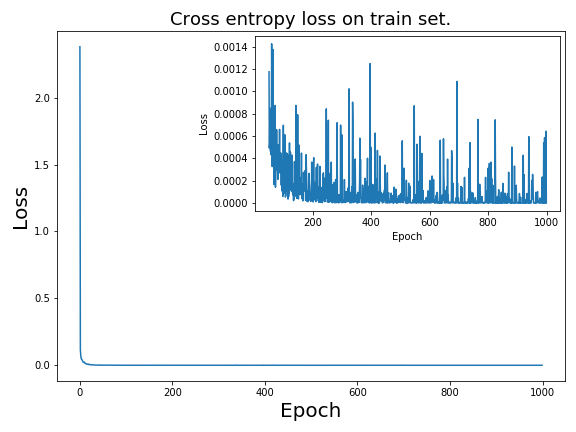}
\includegraphics[width=0.49\textwidth]{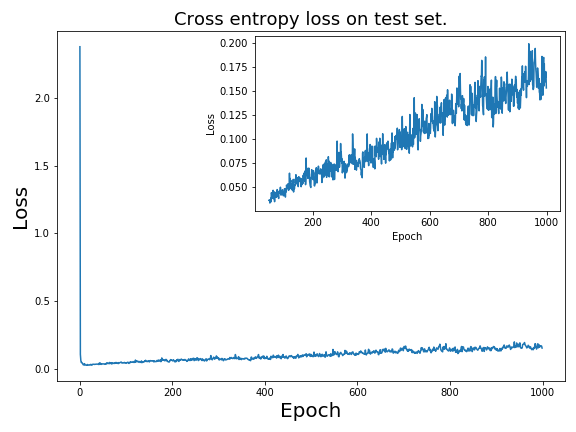}
\includegraphics[width=0.32\textwidth]{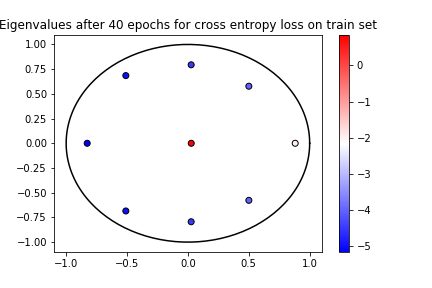}
\includegraphics[width=0.32\textwidth]{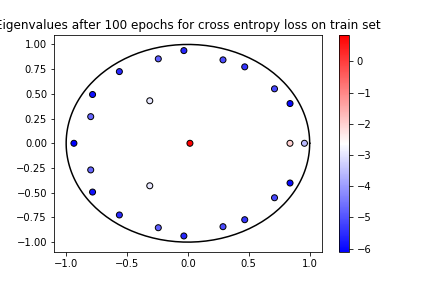}
\includegraphics[width=0.32\textwidth]{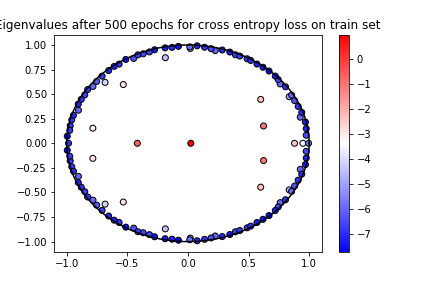}
\includegraphics[width=0.32\textwidth]{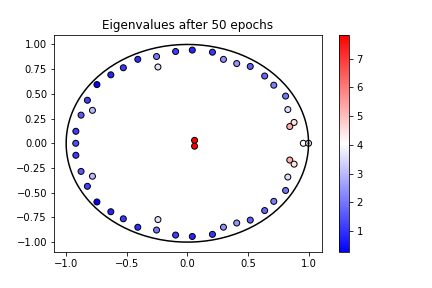}
\includegraphics[width=0.32\textwidth]{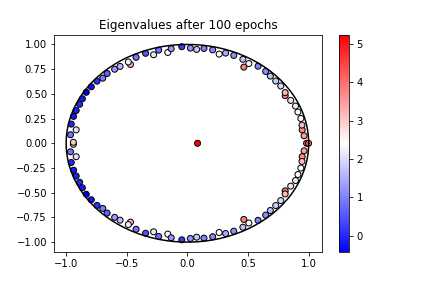}
\includegraphics[width=0.32\textwidth]{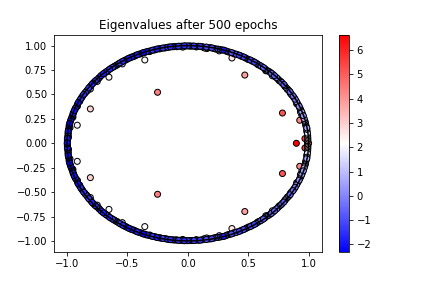}
\caption{\textbf{KMD analysis of network 1 training with Xavier initialization scheme}. \textbf{(1st Row left)} Cross entropy loss of the network on the training set. \textbf{(1st Row right)} Cross entropy loss on the test set. \textbf{(2nd Row)} KMD eigenvalues using the cross entropy on the training set as the observable. Left: computed after 40 epochs. Middle: 100 epochs. Right: 500 epochs. \textbf{(3rd Row)} KMD eigenvalues using the weights vectors during training as the observable. Left: computed after 50 epochs. Middle: 100 epochs. Right: 500 epochs.}
\label{fig:xavier-kmd}
\end{center}
\end{figure}

Figure \ref{fig:xavier-kmd} shows the results for the Xavier initialization scheme. The top left figure shows the cross entropy loss of the network evaluated on the training set at each epoch. The top right figure shows the cross entropy on the test set. The second row shows the KMD spectrum using the cross entropy loss function on the training set as the observable. The spectrum is computed using the first 40 (left), 100 (middle), and 500 (rights) snapshots of the observable. The third row shows the spectrum computed using the weight vectors as the vector-valued observable. The spectrum is computed using the first 50 (left), 100 (middle), and 500 (rights) snapshots of the observable. The spectrum was computed using \textsf{DMD\_RRR}.

Each of the initialization schemes trains very fast, with a large drop in the training error after a few epochs. The HE and Xavier initialization schemes seem to over-memorize the training set which we see as an increase in the cross entropy loss on the test set. However, the final errors on the test set are still lower than the final error for the random normal scheme on the test set.

Since for all initialization schemes, the training process is extremely fast, with a large drop in training set cross entropy, the eigenvalues clustered close to 0 in the second row of the figures (training set cross entropy observable) makes sense. As more snapshots are taken to be used to compute the spectrum, there seems to some important eigenvalues showing up close to 1, which would indicate that the training is nearing a fixed point for the training process. This trend seems clear in the final row of the plots which used the weights vector as the observable for the KMD computation. After 50 epochs, there are important eigenvalues clustered close to 0 and, as more snapshots are taken, important eigenvalues appear and cluster around 1.

\section{Pruning network weights using Koopman modes.}\label{sec:pruning}

Given the sequence of weights $\set{\mat w_t}$, the Koopman mode decomposition can be written as $\mat w_t = \sum_{k} \mat m_k \lambda_k^t \phi_k  + \mat e_t$,
where $\mat m_k$ is the Koopman mode normalized to norm 1, $\lambda_k$ is the associated eigenvalue, $\phi_k$ is the reconstruction coefficient and $\mat e_t$ is the error term at epoch $t$.

The objective here is to use the Koopman mode reconstruction to selectively set weights to zero in order to prune the network. Before each epoch, the weights are collected as snapshots and after a certain number of training epochs, the KMD analysis is performed. Modes are sorted in descending order of absolute value of coefficient in reconstruction of first snapshot; i.e. $\mat m_j \geq \mat m_\ell$ if and only if $\abs{c_j} \geq \abs{c_\ell}$. For reconstruction, we only include the mode corresponding to eigenvalue 1 and any other modes which are ranked higher in the above ordering. This reconstruction is used in algorithm \ref{alg1} or \ref{alg2} to selectively set weights to zero and pinned there. The difference between the two algorithms is in how they handle the weights that were not set to zero. In algorithm \ref{alg1}, the remaining weights are not modified. In algorithm \ref{alg2}, the remaining weights are randomly reinitialized according to the initialization scheme we are using---HE or Xavier.

\RestyleAlgo{boxruled}

\begin{minipage}{0.49\textwidth}
\begin{algorithm}[H]
	\SetAlgoLined
\ \ 	\textbf{Input:} Y:KMD reconstruction of weights \\ 
\ \ $\epsilon$ : threshold for pruning\\
\ \ 	\textbf{Result:} new weights W\\
	
\ \ 	Y[$|Y| < \epsilon$] = 0\\
\ \ 	W = Y (reshaped into layer sizes)\\
\ \ 	\textbf{Return} W
	\caption{Pruning with reconstruction${}_{}$}
	\label{alg1}
\end{algorithm}
\end{minipage}
\hfill
\begin{minipage}{0.49\textwidth}
\begin{algorithm}[H]
	\SetAlgoLined
\ \ 	\textbf{Input:} Y:KMD reconstruction of weights \\ 
\ \ $\epsilon$ : threshold for pruning\\
\ \	\textbf{Result:} new weights W\\
\ \	W = random weights \\	
\ \	mask = [$|Y| < \epsilon$] ; 
	W[mask] = 0 \\
\ \	\textbf{Return} W
	\caption{Random init. with pruning}
	\label{alg2}
\end{algorithm}
\end{minipage}

\begin{figure}[htbp]
\begin{center}
\includegraphics[width=0.49\textwidth]{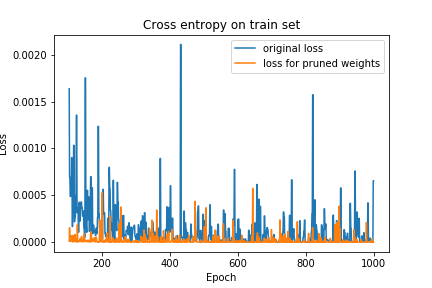}
\includegraphics[width=0.49\textwidth]{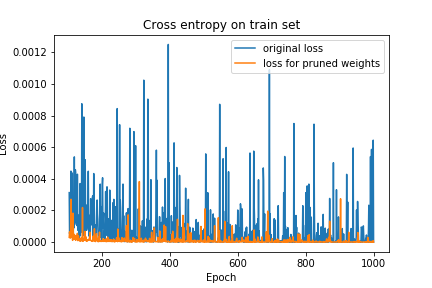}
\includegraphics[width=0.49\textwidth]{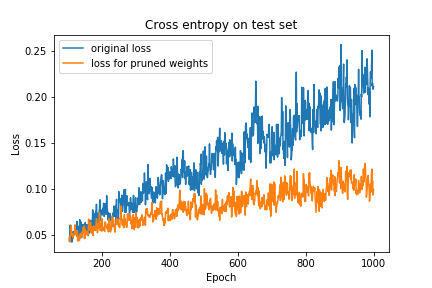}
\includegraphics[width=0.49\textwidth]{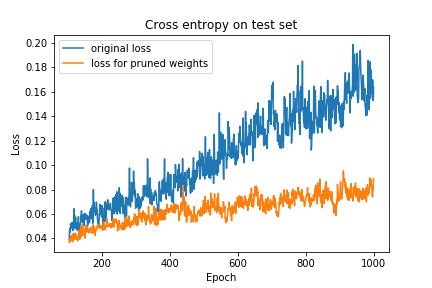}
\caption{\textbf{Pruning network 1 with algorithm 1 at 100 epochs}. Blue is the loss on training set for the unpruned network. Orange is the loss for the pruned network. \textbf{(Top Row)} Cross entropy loss on training set. \textbf{(Bottom Row)} Cross entropy loss on test set. \textbf{(Left column)} HE initialization scheme. \textbf{(Right column)} Xavier initialization scheme. For the HE and Xavier initialization schemes, the pruned network gives comparable results on the training set and better results on the test set as compared to the unpruned network. Additionally, approximately 75\% of the weights have been set to zero in the pruned network.}
\label{fig:alg-1-pruning-100-epochs}
\end{center}
\end{figure}

Pruning the network according to algorithm 1 early in the training process (100 epochs) gives comparable performance to the unpruned network on the training set. For the HE and Xavier initialization schemes, pruning gives better performance of the loss function of the training set (see figure \ref{fig:alg-1-pruning-100-epochs}). In the figures, the blue trace corresponds to the loss on the training set for the unpruned network. The orange trace corresponds to the training set loss on the pruned network. The main takeaway here is that the pruned network have about 75\% of their weights pinned at zero, while still maintaining or beating the performance of the unpruned network. Results for pruning at 500 epochs exhibit the same trend (see Supp. Material). Results for algorithm 2 are similar and are in the Supplemental Material.

\section{Determining the depth of the Hierarchical SVR networks.}\label{sec:hsvr}

In many physics problems, we encounter signals composed of multiple characteristic scales. These multiscale signals often pose problems for machine learning algorithms such as support vector regression (SVR). To mitigate this problem, the authors of \cite{bellocchio2012hierarchical} proposed a hierarchical SVR model 

\begin{equation}
S(x) = \sum_{\ell=0}^{L} a_{\ell}(x; \gamma_\ell)    
\end{equation}
where $a_{\ell}(x; \gamma_\ell)$ is an SVR model \cite{Cortes:1995wa} at the the scale $\gamma_\ell$. Each SVR models use the basis functions $G_\gamma(x; c) = \exp(-\gamma\norm{x-c}^2)$, with the scale functions satisfying $\gamma_{\ell+1} > \gamma_\ell > 0$ so that the first layers are attuned to coarser scale features and subsequent layers are attuned to progressively finer scale features. This architecture is similar to some deep learning architectures, but with different basis functions. But, how does one choose the scales $\gamma_\ell$ and the total number of scales $L$?
\begin{wrapfigure}[18]{o}{0.35\textwidth}
    \centering
        \includegraphics[width=0.35\textwidth]{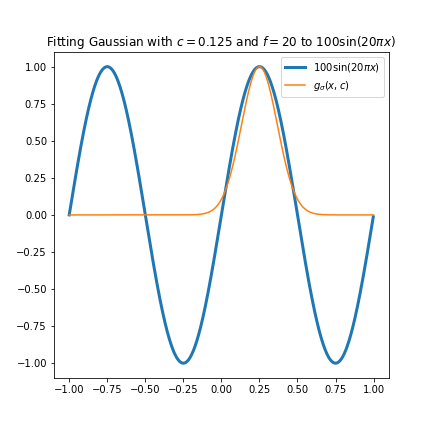}
    \caption{Relating the sinusoid frequency with the SVR spatial model. $\sqrt{1/\gamma} = \sigma = 1/6f$.}
    \label{fig:sin-gaussian}
\end{wrapfigure}

Here, we analyze the spectrum of the signal, using either a Fourier transform or Koopman spectral analysis, to determine the specific scales and how many layers that should be used in the HSVR model. These methods are fast and obviate the need for expensive grid search optimization. We leave the details to the supplemental material and give a high level description of the idea here. We use a Fourier transform or Koopman spectral analysis to compute the frequencies inherent in the data set. These frequencies are ranked from slowest (corresponding to coarse spatial scales) to fastest (corresponding to the finest spatial scale). These frequencies can then be related via an analytical formula to the SVR scales $\gamma_{\ell}$. Instead of taking all the frequencies, because many can be close and offer redundant information, we take the smallest scale $\gamma_0$ and the largest scale scale $\gamma_{\max}$, and \emph{choose} a decay rate, say $\rho=2$, and specify $\gamma_{\ell+1} = \rho \gamma_\ell$. The number of layer required to span the range $[\gamma_0, \gamma_{\max}]$ is computed as $L = \min\set{\ell : \rho^{\ell} \gamma_0 \geq \gamma_{\max}}$. We should note that this procedure also works for single scale signals. The result will be that only a single scale $\gamma_0$ will derived from the spectrum and we get a single layer HSVR model.

More precisely, let us relate the frequency of a sinusoidal signal $s(x) = \sin(2\pi f x)$ to the scale $\gamma$. The scale $\gamma$ is related to the standard deviation, $\sigma$, of the Gaussian as $\gamma = 1/\sigma^2$. To relate the frequency, $f$, of the sinusoid to the scale, $\sigma$, we use the heuristic that we want 3 standard deviations of the Gaussian to be half of the period $T$. That is we want $3\sigma = T/2$. Since $T = 1/f$, we get that $\sigma = 1/6f$. Figure \ref{fig:sin-gaussian} justifies this heuristic. Applied to the signal $h(x) = x + \sin(2\pi x^4)$ on the interval $[0,2]$ and using $\rho = 2$ gives the number of layers as 7.

\section{Noise rejections using negative Sobolev norms in the loss function.}\label{sec:sobolev}

Here, we investigate the performance of different loss functions compared to the standard $L^2$ loss when learning a multiscale signal. The loss functions that we use are inspired by the functional form of negative-index Sobolev norms. For functions $f,h : \T^d \to \R$, where $\T^d$ is the $d$-dimensional torus $(\R/\Z)^d$, the Sobolev norm of order $p=2$ and index $s < 0$ of their difference can be computed via
	\vspace{-1mm}
	\begin{equation}\label{eq:sobolev-norm}
	\norm{f - h}_{H^{s}}^2 = \sum_{\mat k \in \Z^d}  \frac{\abs{\widehat f(\mat k) - \widehat h(\mat k)}^2}{(1 + (2\pi \norm{\mat k}_2)^2)^{-s}},
	\end{equation}
where $\widehat f$ and $\widehat h$ are the Fourier transforms of $f$ and $h$, respectively. As the norm $\norm{\mat k}_2$ of the wave vector increases, the contribution of the term $\abs{\widehat f(\mat k) - \widehat h(\mat k)}^2$ to the loss diminishes; discrepancies between $f$ and $h$ at small scales are not as important as discrepancies at larger/coarser scales.

We use the Sobolev loss functions to fully connected neural networks with $L$ layers $(\ell = 0,\dots, L-1)$ in two ways. Let $\mat h: \R^D \to \R^M$ be the function to be learned and $s : \R \to \R$ be a monotonically increasing function that defines a scale. Usually we will use the linear function $s(\ell) = \ell$ or the exponential function $s(\ell) = 2^\ell$.

\vspace{-3mm}
\paragraph{Sobolev loss 1.} For each hidden layer $\ell$, we define an auxiliary output for that layer denoted by $\mat f_\ell = \mat C_\ell \mat z_\ell$, where $\mat z_\ell \in \R^{N_\ell}$ are the activation functions for layer $\ell$ and $\mat C_\ell \in \R^{M\times N_\ell}$ is a matrix. We add a loss function for each auxiliary output having the form
	\begin{equation}\label{eq:sobolev-loss-1}
	L_\ell( \mat h, \mat f_\ell ) = \sum_{m=1}^{M} \sum_{\norm{\mat k}_1 = s(\ell)} \frac{\abs*{\widehat{\mat h}^{(m)}(\mat k) - \widehat{\mat f}_\ell^{(m)}(\mat k)}^2}{(1 + (2\pi \norm{\mat k}_2)^2)^{1/2}}, \qquad \ell \in  \{1, \dots, L\},
	\end{equation}
where $\widehat{\mat h}^{(m)}$ is the Fourier transform of the $m$-th component function of $\mat h = (h^{(1)},\dots, h^{(M)})$.
By applying a loss of different scales, $s(\ell)$, at each layer, we are enforcing a more interpretable network. Note that for $\ell = 0$, the denominator of \eqref{eq:sobolev-loss-1} is 1 and by Parseval's identity, the expression is equivalent to the $L^2$ norm.

\vspace{-3mm}
\paragraph{Sobolev loss 2.} Here, instead of applying pieces of \eqref{eq:sobolev-norm} at each layer, we apply it only at the output of the final layer, $\mat f_{L-1} =\mat C_{L-1}\mat z_{\ell}$:
\vspace{-3mm}
	\begin{equation}
	L(\mat h, \mat f_{L-1}) = \sum_{m=1}^M \sum_{\ell=0}^{L-1} \sum_{\norm{\mat k}_1 = s(\ell)}  \frac{\abs{\widehat {\mat h}^{(m)}(\mat k) - \widehat {\mat f}_{L-1}^{(m)}(\mat k)}^2}{(1 + (2\pi \norm{\mat k}_2)^2)^{1/2}}.
	\end{equation}
\vspace{-2mm}

\begin{figure}[H]
\begin{center}
\includegraphics[width=0.32\textwidth]{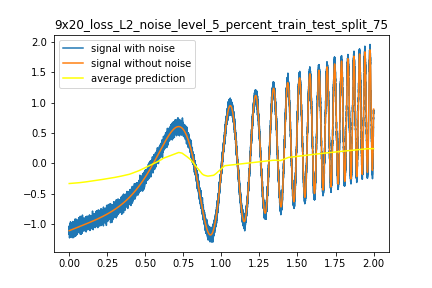}
\includegraphics[width=0.32\textwidth]{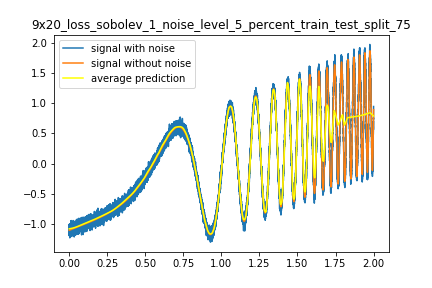}
\includegraphics[width=0.32\textwidth]{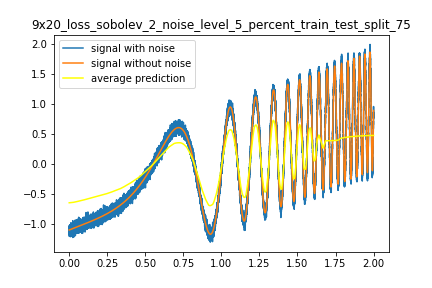}
\includegraphics[width=0.32\textwidth]{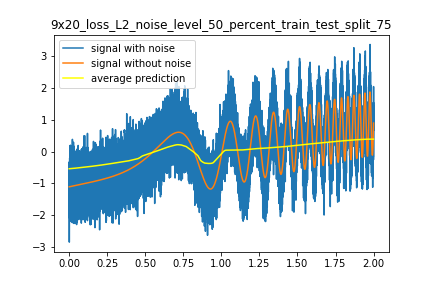}
\includegraphics[width=0.32\textwidth]{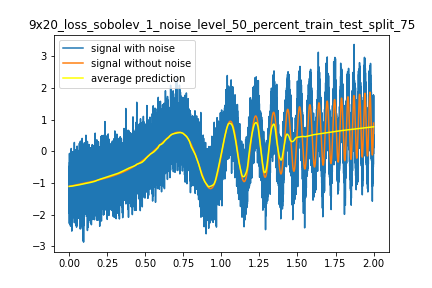}
\includegraphics[width=0.32\textwidth]{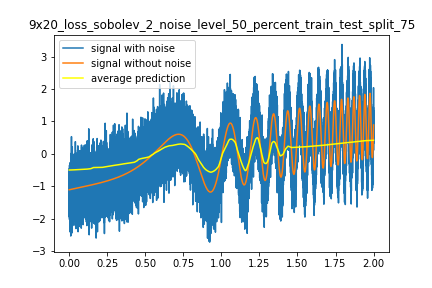}
\includegraphics[width=0.32\textwidth]{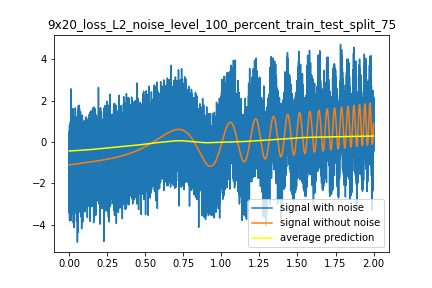}
\includegraphics[width=0.32\textwidth]{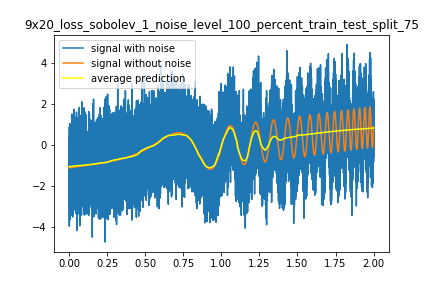}
\includegraphics[width=0.32\textwidth]{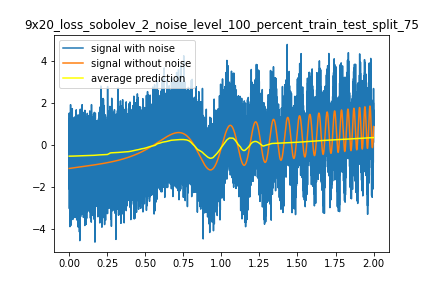}
\caption{Reconstructing a mulitscale signal polluted by noise. \textbf{(Left column)} Pure $L^2$ loss function. \textbf{(Middle column)} Sobolev loss 1. \textbf{(Right column)} Sobolev loss 2. Each row, top to bottom, corresponds to noise levels $\epsilon = 0.05, 0.5$, and $1.0$, respectively. The noisy signal $\widehat h$ was used as the data set with the goal of reconstructing the clean signal $h$. A train/test split of 75/25 was used on $\set{\widehat h(x)}$.}
\label{fig:sobolev-loss}
\end{center}
\end{figure}

We apply the two methods on the multiscale signal $h(x) = x + \sin(2\pi x^4)$ (therefore $M$=1) on the interval $[0,2]$. At each point $x$, we add noise $\eta(x)$ to the signal that is distributed according to 
	\begin{equation}
	\eta \sim \epsilon \left(\max_{x\in[0,2]} h(x) - \min_{x\in[0,2]} h(x)\right) N(0,1).
	\end{equation}
The noisy signal $\widehat h = h + \eta$ is used as the data set. Figure \ref{fig:sobolev-loss} shows the performance of the different loss functions using the noisy signal to train. The left column contains the result for pure $L^2$ loss function, the middle column is Sobolev loss 1, and the third column is Sobolev loss 2. Each row, top to bottom, corresponds to noise levels $\epsilon = 0.05, 0.5$, and $1.0$, respectively. The network had 9 hidden layers each containing 20 neurons and a single node output layer. The choice of 9 was informed by the HSVR analysis above so we would have slightly more layers than the minimum specified by the HSVR analysis. Layers were fully-connected. As can be seen the pure $L^2$ loss function reconstructs the clean signal $h$ poorly at every noise level, basically reconstructing the average of the signal. Of the two Sobolev loss functions, the first performs the best at reconstructing the clean signal, even in the presence of high amounts of noise. This is likely due to imposing a specific scale for each layer, rather than trying to force the network to try to disentangle the scales at the final layer. For each of the Sobolev loss, the linear scale function, $s(\ell) = \ell$ was used.

\section{Conclusions}
In this paper, we have introduced results on using the Koopman operator to analyze aspects of neural networks. In order to do this, we consider the training process as a dynamical system on the weights. In particular, we used the spectrum of the Koopman operator to determine when to terminate training, used the Koopman modes to prune a networks weights with the pruned network having similar performance to he unpruned network. Additionally, we introduced a method based on either the Fourier or Koopman spectrum to determine the required number of layer in a Hierarchical SVR model for a mulitscale signal. These same ideas apply to using neural networks to model such signals. Finally, we introduced loss functions based on negative Sobolev norms which allow significant noise rejection when trained on a noisy multimodal signal.


\section{Acknowledgment}
This work was partially supported under DARPA contract HR0011-18-9-0033.

\bibliographystyle{plain}
\bibliography{bib/refs}


\input{supplemental-for-include.tex}

\end{document}

%% file: supplemental-for-include.tex
\section*{Supplemental Material}

\section{Insights into neural network training convergence via the Koopman spectrum.}

The standard MNIST data set was used for testing the methods. The network tested was a convolutional network with two convolutional layers with $16$ and $32$ kernels, each followed by $2 \times 2$ max-pooling. Each kernel is $5 \times 5$. Convolutional layers are followed by fully-connected layer with $100$ neurons, which is then followed by another fully-connected layer with $10$ neurons, after which follows softmax classification. The architecture is shown on Figure \ref{fig:small-nn-s}. A cross-entropy loss function was used with a learning rate of $\eta = 1e$-3. Different weight initialization (He, Xavier, or standard normal) were tested as well. After each epoch, cross-entropy losses on the train and test sets were recorded along with the network weights. All networks were trained for $1000$ epochs, and KMD analysis was applied to the snapshots for 3 different epoch ranges specified in the figures. After that, diverging models were trained for additional $1000$ epochs, and diverging behaviour continued. 
\setcounter{figure}{0}
\begin{figure}[htbp]
\begin{center}
\includegraphics[width=0.9\textwidth]{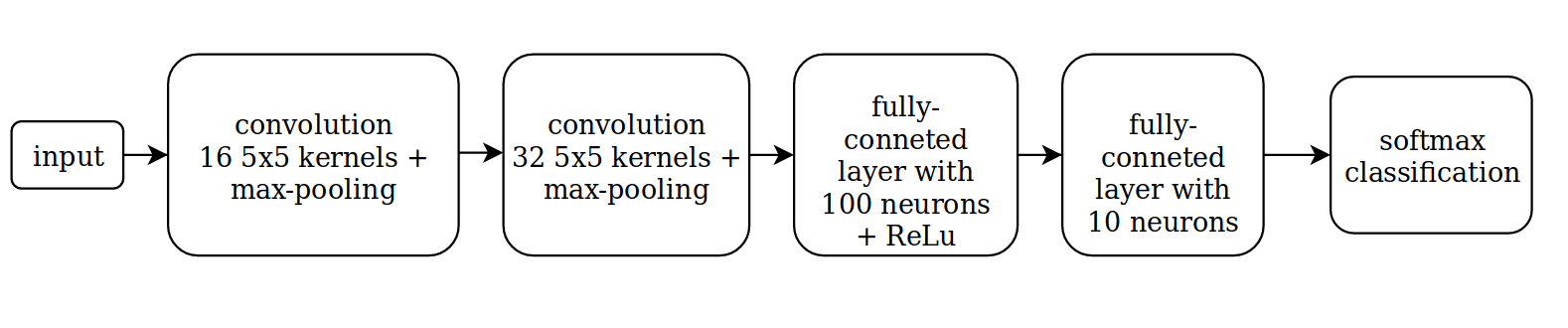}
\caption{Network architecture.}
\label{fig:small-nn-s}
\end{center}
\end{figure}

%

\begin{figure}[htbp]
\begin{center}
\includegraphics[width=0.49\textwidth]{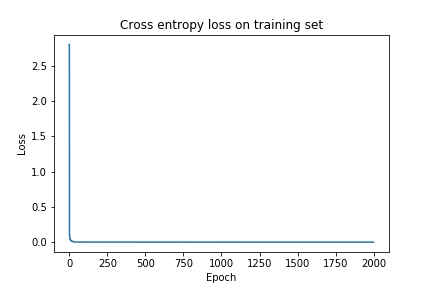}
\includegraphics[width=0.49\textwidth]{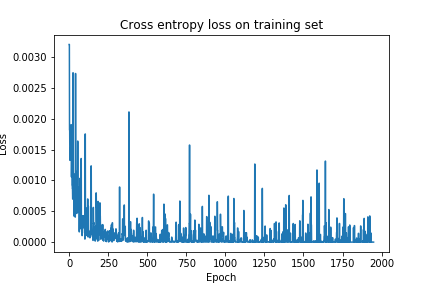}
\includegraphics[width=0.49\textwidth]{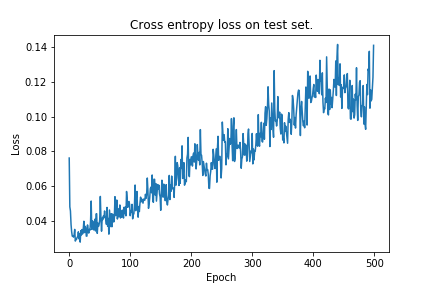}
\includegraphics[width=0.49\textwidth]{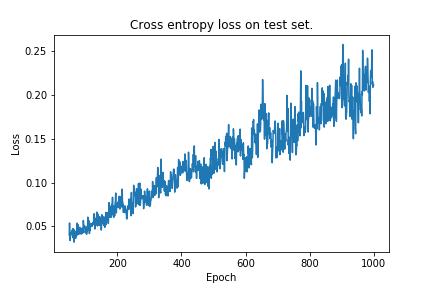}
\includegraphics[width=0.32\textwidth]{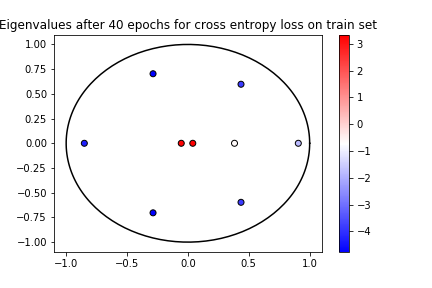}
\includegraphics[width=0.32\textwidth]{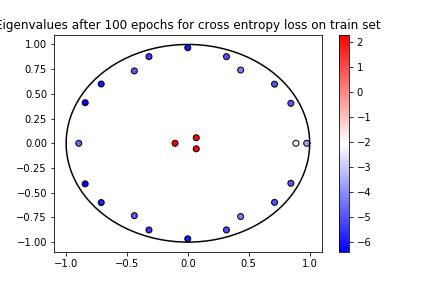}
\includegraphics[width=0.32\textwidth]{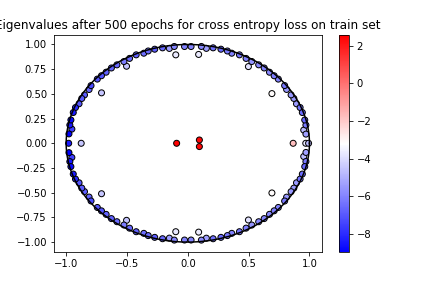}
\includegraphics[width=0.32\textwidth]{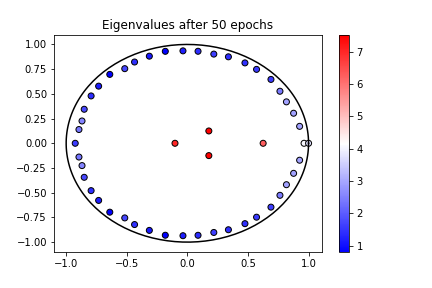}
\includegraphics[width=0.32\textwidth]{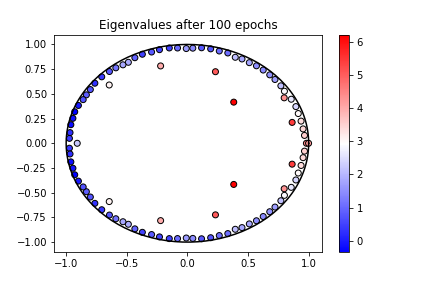}
\includegraphics[width=0.32\textwidth]{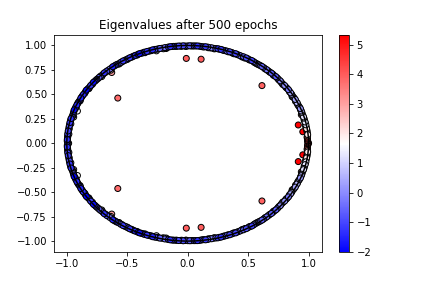}
\caption{\textbf{KMD analysis of network 1 training with HE initialization scheme}. \textbf{(1st Row)} Cross entropy loss of the network on the training set. Left: loss from epoch 0 onward. Right: loss from epoch 50 onward. \textbf{(2nd Row)} Cross entropy loss on the test set. Left: loss from epoch 0 onward. Right: loss from epoch 50 onward. \textbf{(3rd Row)} KMD eigenvalues using the cross entropy as the observable. Left: computed after 40 epochs. Middle: 100 epochs. Right: 500 epochs. \textbf{(4th Row)} KMD eigenvalues using the weights vector as the observable. Left: computed after 50 epochs. Middle: 100 epochs. Right: 500 epochs.}
\label{fig:he-kmd}
\end{center}
\end{figure}
\begin{figure}[htbp]
\begin{center}
\includegraphics[width=0.49\textwidth]{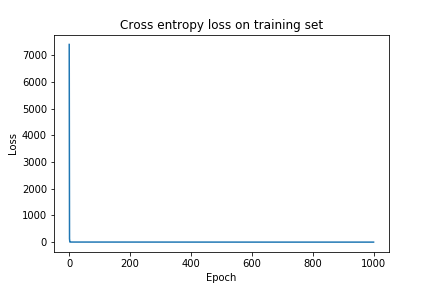}
\includegraphics[width=0.49\textwidth]{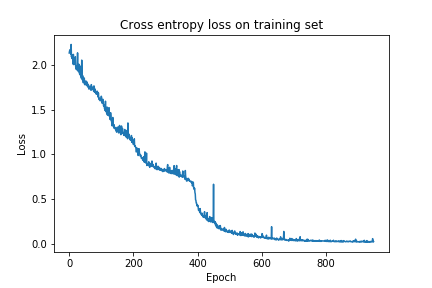}
\includegraphics[width=0.49\textwidth]{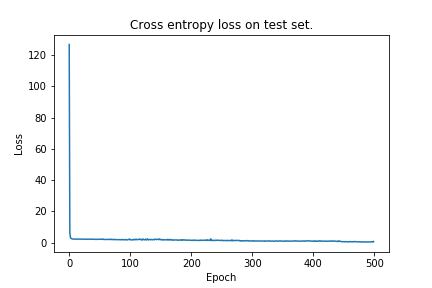}
\includegraphics[width=0.49\textwidth]{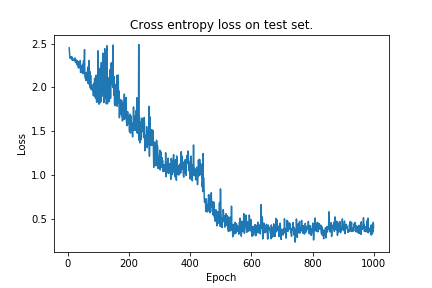}
\includegraphics[width=0.32\textwidth]{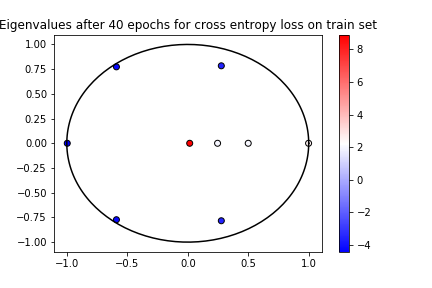}
\includegraphics[width=0.32\textwidth]{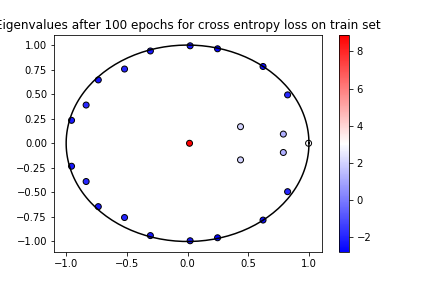}
\includegraphics[width=0.32\textwidth]{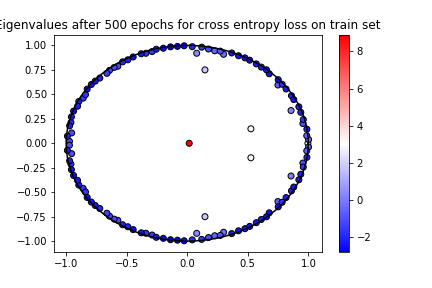}
\includegraphics[width=0.32\textwidth]{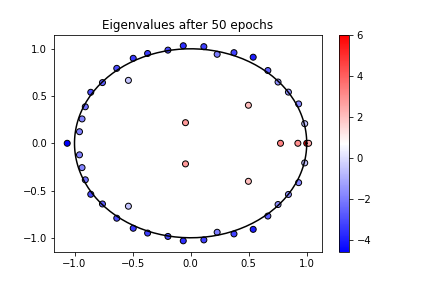}
\includegraphics[width=0.32\textwidth]{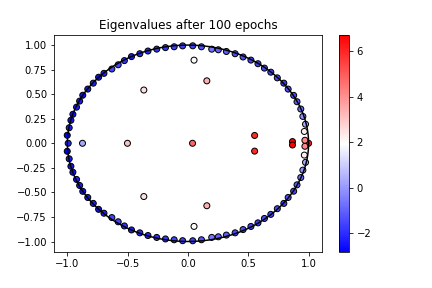}
\includegraphics[width=0.32\textwidth]{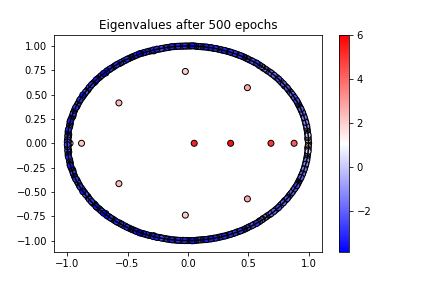}
\caption{\textbf{KMD analysis of network 1 training with random normal initialization scheme}. \textbf{(1st Row)} Cross entropy loss of the network on the training set. Left: loss from epoch 0 onward. Right: loss from epoch 50 onward. \textbf{(2nd Row)} Cross entropy loss on the test set. Left: loss from epoch 0 onward. Right: loss from epoch 50 onward. \textbf{(3rd Row)} KMD eigenvalues using the cross entropy as the observable. Left: computed after 40 epochs. Middle: 100 epochs. Right: 500 epochs. \textbf{(4th Row)} KMD eigenvalues using the weights vector as the observable. Left: computed after 50 epochs. Middle: 100 epochs. Right: 500 epochs.}
\label{fig:rn-kmd}
\end{center}
\end{figure}

Figures \ref{fig:he-kmd} and \ref{fig:rn-kmd} hold the results for the HE and random normal initialization schemes, respectively. Results for the Xavier initialization scheme were presented in the main text. The top row of each figure shows the cross entropy loss of the network evaluated on the training set at each epoch. The second row shows the cross entropy on the test set. For each of these two rows, the left figure shows the cross entropy from epoch 0 onward, the right figure from epoch 50 onward to show a zoomed in view. The third row shows the KMD spectrum using the cross entropy loss function on the training set as the observable. The spectrum is computed using the first 40 (left), 100 (middle), and 500 (rights) snapshots of the observable. The fourth row show the spectrum computed using the weight vectors as the vector-valued observable. The spectrum is computed using the first 50 (left), 100 (middle), and 500 (rights) snapshots of the observable. The spectrum was computed using \textsf{DMD\_RRR}.


The conclusions for these initialization schemes remain the same as for the Xavier initialization scheme. For all initialization schemes, the training process is extremely fast, with a large drop in training set cross entropy, the eigenvalues clustered close to 0 in the third rows of the figures (training set cross entropy observable) makes sense. As more snapshots are taken to be used to compute the spectrum, there seems to be some important eigenvalues showing up close to 1, which would indicate that the training is nearing a fixed point for the training process. This trend seems clear in the final row of the plots which used the weights vector as the observable for the KMD computation. After 50 epochs, there are important eigenvalues clustered close to 0 and, as more snapshots are taken, important eigenvalues appear and cluster around 1.

\clearpage
\section{Pruning network weights using Koopman modes.}

Given the sequence of weights $\set{\mat w_t}$, the Koopman mode decomposition (see main text) can be written as $\mat w_t = \sum_{k} c_k \lambda_k^t \mat m_k + \mat e_t$,
where $\mat m_k$ is the Koopman mode normalized to norm 1, $\lambda_k$ is the associated eigenvalue, $c_k$ is the reconstruction coefficient and $\mat e_t$ is the error term at epoch $t$.



\setcounter{algocf}{0}
\begin{minipage}{0.47\textwidth}
\begin{algorithm}[H]
	\SetAlgoLined
	\textbf{Input:} Y:KMD reconstruction of weights , \\$\epsilon$ : threshold for pruning\\
	\textbf{Result:} new weights W\\
	
	Y[$|Y| < \epsilon$] = 0\\
	W = Y (reshaped into layer sizes)\\
	\textbf{Return} W
	\caption{Pruning with reconstruction${}_{}$}
	\label{alg1-s}
\end{algorithm}
\end{minipage}
\hfill
\begin{minipage}{0.47\textwidth}
\begin{algorithm}[H]
	\SetAlgoLined
	\textbf{Input:} Y:KMD reconstruction of weights , \\$\epsilon$ : threshold for pruning\\
	\textbf{Result:} new weights W\\
	W = random weights \\	
	mask = [$|Y| < \epsilon$] ; 
	W[mask] = 0 \\
	\textbf{Return} W
	\caption{Random initialization with pruning}
	\label{alg2-s}
\end{algorithm}
\end{minipage}

\begin{figure}[htbp]
\begin{center}
\includegraphics[width=0.48\textwidth]{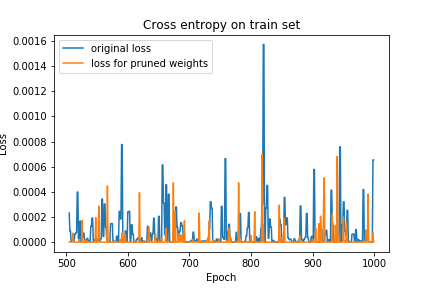}
\includegraphics[width=0.48\textwidth]{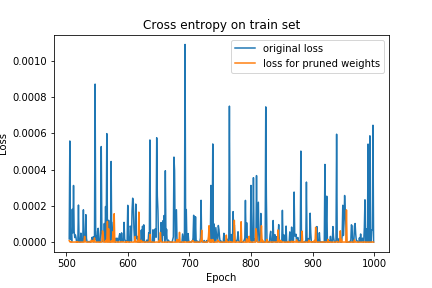}
\includegraphics[width=0.48\textwidth]{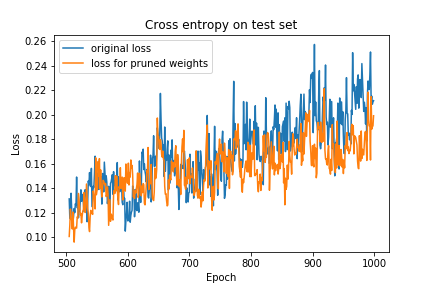}
\includegraphics[width=0.48\textwidth]{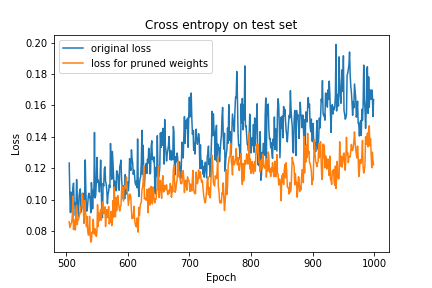}
\caption{\textbf{Pruning the network with Algorithm 1 at 500 epochs}. Blue is the loss on training set for the unpruned network. Orange is the loss for the pruned network. \textbf{(Top Row)} Cross entropy loss on training set. \textbf{(Bottom Row)} Cross entropy loss on test set. \textbf{(Left column)} HE initialization scheme. \textbf{(Right column)} Xavier initialization scheme.}
\label{fig:alg-1-pruning-500-epochs}
\end{center}
\end{figure}

Using Algorithm \ref{alg1-s} to prune at 500 epochs (pruning at 100 epochs is in the main text), the training performance is comparable to the unpruned network for both the HE and Xavier initialization schemes (upon more training after the pruning step). See Figure \ref{fig:alg-1-pruning-500-epochs}.

The results for algorithm 2 are comparable to algorithm 1, for pruning either at 100 or 500 epochs. The pruned networks maintain or beat the unpruned networks while having significantly less weights.

\begin{figure}[htbp]
\begin{center}
\includegraphics[width=0.48\textwidth]{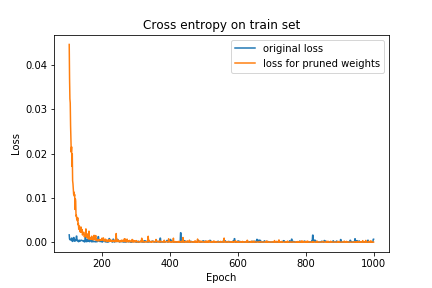}
\includegraphics[width=0.48\textwidth]{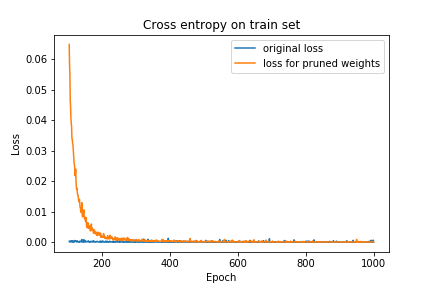}
\includegraphics[width=0.48\textwidth]{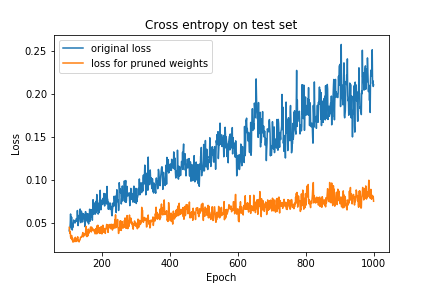}
\includegraphics[width=0.48\textwidth]{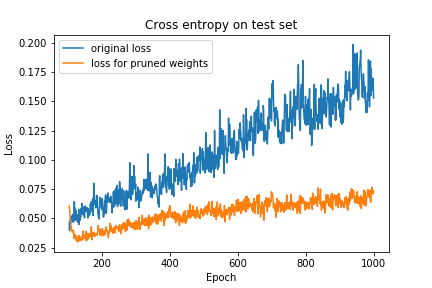}
\caption{\textbf{Pruning the network with Algorithm 2 at 100 epochs}. Blue is the loss on training set for the unpruned network. Orange is the loss for the pruned network. \textbf{(Top Row)} Cross entropy loss on training set. \textbf{(Bottom Row)} Cross entropy loss on test set. \textbf{(Left Column)} HE initialization scheme. \textbf{(Right Column)} Xavier initialization scheme.}
\label{fig:alg-2-pruning-100-epochs}
\end{center}
\end{figure}

\begin{figure}[htbp]
\begin{center}
\includegraphics[width=0.48\textwidth]{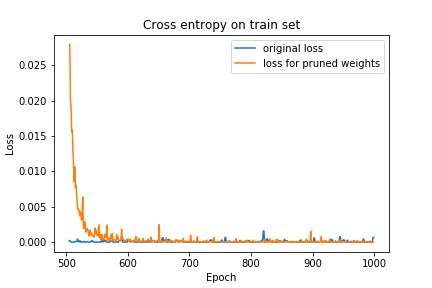}
\includegraphics[width=0.48\textwidth]{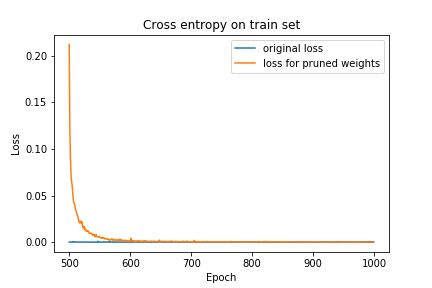}
\includegraphics[width=0.48\textwidth]{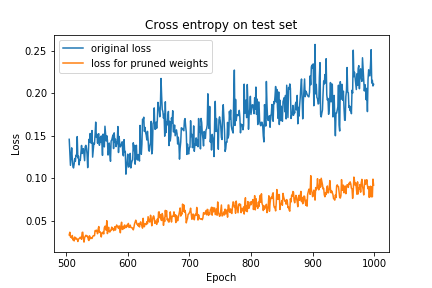}
\includegraphics[width=0.48\textwidth]{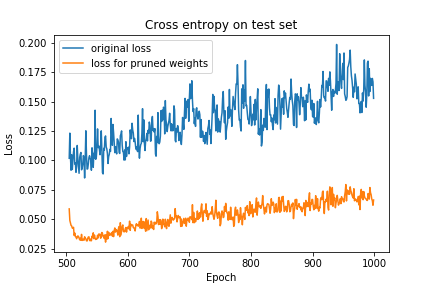}
\caption{\textbf{Pruning the network with algorithm 2 at 500 epochs}. Blue is the loss on training set for the unpruned network. Orange is the loss for the pruned network. \textbf{(Top Row)} Cross entropy loss on training set. \textbf{(Bottom Row)} Cross entropy loss on test set. \textbf{(Left Column)} HE initialization scheme. \textbf{(Right Column)} Xavier initialization scheme.}
\label{fig:alg-2-pruning-500-epochs}
\end{center}
\end{figure}

\section{Determining the depth of the Hierarchical SVR networks.}

Using the methods in the main text (see Sec. 5), Table \ref{table:hsvr-exact} shows the predicted number of layers using both the spectrum of the Fourier transform and DMD for Hierarchical SVR models. Each model is an $\epsilon$-SVR model where $\epsilon$ is a parameter which specifies that if the model and data are $\epsilon$ close, it does not contribute to the overall loss during learning. This implies that the model error cannot decrease below $\epsilon$. The methods were tested for a variety of different functions, both single- and multi-scale, evaluated on $[0,2]$. The number of layers computed by using the spectrums of FFT and DMD are shown along with HSVR models' errors. For each function, the model errors were comparable to $\epsilon$, with some mismatch on the number of layers computed by FFT and DMD. This is likely due to the different methods computing different spectrums and thus number of layers.

\begin{table}[h]
\caption{Predicting the number of HSVR layers for explicitly defined functions. Scales are determined from FFT with decay $\rho=2$ and $\epsilon = 0.01(\max(y)-\min(y))$.} 
	\scalebox{1.0}{
 \begin{tabular}{|cccccc|}
	\hline
	function  & \makecell{nbr of layers\\(FFT)} & $\epsilon$ & \makecell{error\\(FFT)} & \makecell{nbr of layers\\(DMD)} & \makecell{error\\(DMD)} \\ 
	\hline\hline
	$\sin(2\pi x)$ &1 &0.02& 0.02 &1  & 0.02 \\ \hline
	$\sin(20\pi x)$ & 1  & 0.0199 &0.021 &1 & 0.02  \\ \hline
	$\sin(200\pi x)$ &1 & 0.019 & 0.093 & 1 & 0.097 \\  \hline
	$100\sin(20\pi x)$ &1	 & 1.99 & 2 & 1&2.01  \\  \hline
	$40 \cos(2 \pi x)$&1 &0.8 &0.8 & 1 & 0.8 \\  \hline
	$100 \cos(20 \pi x)$ &1 &2 &2.03 &1 &2\\  \hline
	$\sin(2 \pi x^2)$ & 5 &0.0199& 0.02 &1 &0.02  \\  \hline
	$x + x ^2 + x^3$ &2 &0.14 &0.14 &1 & 8 \\ \hline
	$x+\sin(2 \pi x^4)$ & 7 & 0.03 & 0.037 &1 &0.034\\ \hline
	$\cos(2 \pi x) + \sin(20 \pi x)$  & 2 & 0.0397 & 0.0404 &2 &0.042\\ \hline
	$\cos(20 \pi x) \sin(15 \pi x)$ & 2 & 0.02 & 0.021 & 2 &0.022\\ \hline
	$\cos(32 \pi x)^3$ & 1 & 0.0199 & 0.022 & 2& 0.022\\ \hline
\makecell{	$\sin(13 \pi x) + \sin(17 \pi x) +$\\$ \sin(19 \pi x) + \sin(23 \pi x)$} & 1 & 0.076 & 0.077 & 1&0.077 \\ \hline	
	$ \sin(50 \pi x) \sin(20 \pi x) \cos(15 \pi x)$ &3 &0.0187 & 0.02 & 2 &0.02\\ \hline
	\makecell{$\sin(40 \pi x) \cos(10 \pi x) +$\\$ 3\sin(20 x)\sin(40x)$} &5 &0.064 & 0.065 &3 & 0.066\\ \hline
	$\sin(2x)\cos(32x)$ & 5  & 0.0198 & 0.02 & 1&0.02 \\
	\hline
\end{tabular}}
\label{table:hsvr-exact}
\end{table}